\documentclass[10pt, conference, compsocconf]{IEEEtran}
\usepackage{ifpdf}
\usepackage{cite}
\ifCLASSINFOpdf
   \usepackage[pdftex]{graphicx}
   \graphicspath{{./Images/}}
   \DeclareGraphicsExtensions{.pdf,.jpeg,.png}
\else
   \usepackage[dvips]{graphicx}
   \DeclareGraphicsExtensions{.eps}
\fi
\usepackage[cmex10]{amsmath}

\usepackage{array}

\usepackage{eqparbox}

\usepackage[tight,footnotesize]{subfigure}
\usepackage[font=footnotesize]{caption}
\usepackage[font=footnotesize]{subfig}
\usepackage{stfloats}

\usepackage{url}

\usepackage{color}
\usepackage{mathptmx} 
\usepackage{times} 
\usepackage{amsmath} 
\usepackage{amssymb}  
\usepackage{tikz}
\usetikzlibrary{shapes.geometric, arrows}
\usepackage{algorithm}
\usepackage[noend]{algpseudocode}

\tikzstyle{startstop} = [rectangle, rounded corners, minimum width=3cm, minimum height=1cm,text centered, draw=black]
\tikzstyle{io} = [trapezium, trapezium left angle=70, trapezium right angle=110, minimum width=3cm, minimum height=1cm, text centered, draw=black ]
\tikzstyle{process} = [rectangle, minimum width=3cm, minimum height=1cm, text centered, draw=black]
\tikzstyle{decision} = [diamond, minimum width=3cm, minimum height=1cm, text centered, draw=black]
\tikzstyle{arrow} = [thick,->,>=stealth]

\begin{document}
%
\title{SLAM-Assisted Coverage Path Planning for Indoor LiDAR Mapping Systems}


\author{\IEEEauthorblockN{Ankit Manerikar}
\IEEEauthorblockA{School of Electrical and Computer Engg.\\
Purdue University\\
West Lafayette, USA.\\
Email: amanerik@purdue.edu}
\and
\IEEEauthorblockN{Tamer Shamseldin}
\IEEEauthorblockA{Lyles School of Civil Engineering\\
Purdue University\\
West Lafayette, USA.\\
Email: tshamsel@purdue.edu}
\and
\IEEEauthorblockN{Ayman Habib}
\IEEEauthorblockA{Lyles School of Civil Engineering\\
Purdue University\\
West Lafayette, USA.\\
Email: ahabib@purdue.edu}
}


%


\maketitle

\begin{abstract}
Applications involving autonomous navigation and planning of mobile agents can benefit greatly by employing online Simultaneous Localization and Mapping (SLAM) techniques, however, their proper implementation still warrants an efficient amalgamation with any offline path planning method that may be used for the particular application. In this paper, such a case of amalgamation is considered for a LiDAR-based indoor mapping system which presents itself as a 2D coverage path planning problem implemented along with online SLAM. This paper shows how classic offline Coverage Path Planning (CPP) can be altered for use with online SLAM by proposing two modifications: (i) performing convex decomposition of the polygonal coverage area to allow for an arbitrary choice of an initial point while still tracing the shortest coverage path and (ii) using a new approach to stitch together the different cells within the polygonal area to form a continuous coverage path. Furthermore, an alteration to the SLAM operation to suit the coverage path planning strategy is also made that evaluates navigation errors in terms of an area coverage cost function. The implementation results show how the combination of the two modified offline and online planning strategies allow for an improvement in the total area coverage by the mapping system - the modification thus presents an approach for modifying offline and online navigation strategies for robust operation. 
\end{abstract}


%
\IEEEpeerreviewmaketitle

\section{Introduction}
Coverage Path Planning (CPP) strategies addressing area coverage applications pose as a disparate class of motion planning methods from generic motion planning techniques despite encompassing a large number of applications ranging from aerial surveying and robotic demining \cite{c1} \cite{c2} to autonomous lawn mowing and vacuum cleaning \cite{c3} \cite{c4}. This distinction of coverage path planning can be highlighted by the definition provided by Galceran and Carreras \cite{c5}. Thus, while a general motion planning problem is concerned with generating a cost-optimal trajectory between a source and destination, a CPP problem focuses on generating a path that maximizes area coverage. Therefore, it is only natural to consider that when SLAM techniques need to be employed to ameliorate the performance of a coverage path planner, the former as well as the latter need to be modified to maximize area coverage which is the main goal of the planning action.

The implementation described in this paper has been motivated by our research on developing SLAM-assisted virtual GNSSS/INS frameworks for autonomous operation of Mobile Mapping Systems (MMS) \cite{c6} \cite{c7} for photogrammetry in GNSS-denied environments. Development of such SLAM-based systems for autonomous vehicles has necessitated the use of a hybrid path-planning strategy that combines offline and online planners for complete operation. Accordingly, this paper considers the CPP problem for a 2D LiDAR mapping system designed to map all the points within a bounded target region in finite time. While an offline CPP method can allow the robot to trace a pre-planned path for area coverage, it cannot account for dynamic area coverage corrections during path traversal even with online SLAM. This problem is addressed in the implementation in two stages - first,  the offline CPP strategy is altered to allow for an arbitrary choice of an initial point for the coverage path. Secondly, an area coverage function is evaluated in the online SLAM method to parametrize and correct robot motion for maximizing the area covered.

The paper is organized to explain the mentioned modifications in three stages. First, a brief review of relevant literature to CPP techniques and SLAM is made in Section II followed by an introduction of the concept of SLAM-assisted Coverage Path Planning in Section III. This concept is elaborated by an explanation of the modified offline and online planning strategies in Section IV and V, respectively. Finally, the implementation and results of the proposed methods are illustrated in Section VI and VII that allow a comparative analysis of the modified coverage operation.     
\smallskip

\section{RELATED WORK}

   \begin{figure*}[!h]
      \centering
      \includegraphics[width=17cm]{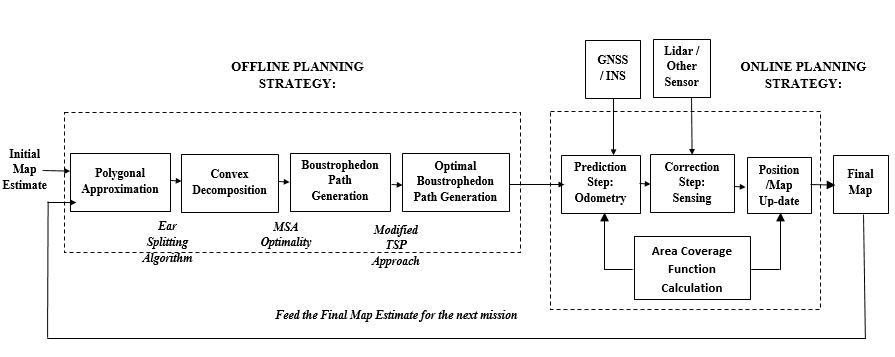}
      \caption{Implementational Block Diagram for SLAM-Assisted Coverage Path-Planning}
      \label{figurelabel}
   \end{figure*}

Since Coverage Path Planning forms a basic framework for many path planning problems, it is addressed in literature for such applications as robotic vacuum cleaning \cite{c4}, autonomous harvesting \cite{c3} and underwater/aerial surveying \cite{c5}. While the basic requirements and constraints for a coverage operation have been aptly defined in the early works on CPP \cite{c9}, a mathematical treatment on the topic by Arkin and Hassin \cite{c10} treats the CPP problem as a variant of the covering salesman problem, which is an NP-hard problem and therefore difficult to optimize without specifying operational constraints. Among the many approaches to solving the CPP problem, this paper focuses on exact coverage planners that ensure complete coverage of the target area with energy/time-optimality. Choset and Pignon \cite{c11} in 2001, proposed the Boustrophedon Cellular Decomposition (BCD) method as a successful exact offline CPP method for complete coverage of a bounded area. However, while Boustrophedon patterns used in this technique allow for complete area coverage, the method does not guarrantee distance/energy optimality for the total path traced by the robot. The Minimum-Sum-of-Altitudes (MSA) decomposition described by Huang \cite{c12} provided a cost function that can be used to optimize this constraint. This criterion has been implemented in UAV path planning problems \cite{c1} to show that minimizing the number of turns can allow for path optimality. It is with regard to this criterion that the offline CPP method is proposed in this paper to allow optimality for an arbitrary choice of an initial point for the path as well as trace the shortest total path for the same. 

The advent of SLAM techniques \cite{c13} \cite{c14} necessitated further modifications to the CPP problem - while the accurate pose and map estimation provided by SLAM can be utilized conveniently for a navigation problem, for a CPP planner, the map/position update from SLAM needs to influence the robot controls in order to maximize area coverage. SLAM implementations for coverage applications have addressed such issues as dynamic coverage planning \cite{c15} and optimal path tracing \cite{c16}. In this paper, we propose a new hybrid strategy that uses an area coverage function allowing dynamic changes to the coverage path from any map/pose updates provided by online SLAM.

\section{SLAM-ASSISTED COVERAGE PLANNING}
This section provides a functional description of different blocks of the implemented LiDAR mapping system which is illustrated in Figure 1. The block diagram demonstrates the system bifurcation into an offline Coverage Path Planner that generates the motion path for navigation, and an online SLAM system uses control input from this path to update the same for dynamic changes during path traversal. The block diagram is briefly described ahead while the theory behind each of the blocks is elaborated in latter sections:
 
\subsection{Block Description:}

\subsubsection{Offline Planning}
The input to the mapping system is an initial coverage map estimate, $M_{init}$, which can be dynamically updated for iterative operations as follows: 
\smallskip
\\
\textit{Polygonal Approximation} -
The proposed CPP planner requires a polygonal approximation of the area boundary for complete coverage - hence,  $M_{init}$ is fed as input in the form of counter-clockwise vertex coordinates of the polygonal approximation of the coverage area. The starting point for the coverage path is also provided. If the provided map is not in polygonal form, it is converted to the same using Douglas-Peucker Polyline Simplification \cite{c17} (a tolerance of 10 cm has been used here).
\smallskip
\\
\textit{Convex Decomposition} -
From the polygonal area provided, the first step is to partition the entire area into a disjoint set of convex cells. This is done through convex decomposition by polygon triangulation using the Ear-Splitting Algorithm. The criterion for convex decomposition involves factors that maintain coverage path optimality (MSA) within each cell.
\smallskip
\\
\textit{Boustrophedon Pattern Generation} -
A Boustrophedon coverage pattern is then generated for each of these convex cells - optimality of coverage in terms of path length is achieved by using the MSA criterion for selecting the optimal sweep direction.
\smallskip
\\
\textit{Optimal Total Path Generation} - 
A continuous path connecting the separate polygon cells with the smallest length of jumps is then generated using a modified Traveling Salesman Problem (TSP) approach that stitches together all the individual paths generated in the previous block more efficiently than sequentially tying them together.
\smallskip

\subsubsection{Online Planning}
The motion path generated by the offline planner is fed to the online system which uses these control inputs for SLAM operation as explained ahead. For this implementation, a GMapping SLAM strategy \cite{c18} is used for online operation using 2D LiDARs as sensors and occupancy grids \cite{c19} for localization and mapping.
\smallskip
\\
\textit{Prediction Step (SLAM)} -
The prediction step for the system is identical to a generic SLAM prediction step - the  coverage path generated by the previous block feeds control inputs to the robot for path traversal. As these inputs are incrementally fed to the robot, the robot pose estimate is updated accordingly.
\smallskip
\\ 
\textit{Correction Step (SLAM)} -
The correction step for SLAM uses the LiDAR point clouds to update the occupancy grid for the environment map - the occupancy grid also plays a crucial role in evaluating the area coverage function that alters the motion of the robot for maximizing area coverage.
\smallskip
\\
\textit{Position / Map Update} -
The current robot position as well as the occupancy grid map are updated continuously while robot traces the entire coverage path. This operation is carried out upon two main events - when the robot trajectory is observed to deviate from the coverage path or if the occupancy grid observes a new dynamic change in the map dimensions requiring a new coverage path to be generated from the offline planner for the new area.
\smallskip
\\
\textit{Area Coverage Function Evaluation} -
The area coverage function is a measure of the total area currently covered with respect to the total target covered area. This is used to control robot motion in two ways. A shape matching operation is carried out between the current occupancy grid and the target area to give a shape error vector which if exceeded beyond a certain threshold  triggers a new coverage path to be created by the offline planner. Secondly, the position update is used to evaluate the current area coverage to provide an optimal direction for the next robot step  that maximizes area coverage while reducing overlap.

\section{OFFLINE PLANNING - CPP STRATEGY}
\smallskip

This section describes in detail the blocks implemented in the offline planner as well as how they are modified for efficient operation.  As this offline planner focuses on the Coverage Path Planning strategy, a comprehensive definition of CPP is first made as given ahead and is followed upon throughout the paper:
\smallskip

\textit{Having given a bounded area to be covered, CPP is the process of generating a finite trajectory such that a robot with a fixed non-zero coverage radius is capable of completely or approximately covering the given area in finite time by tracing the generated trajectory.}
\smallskip

Along these lines, the system in Figure 1 implements an exact cellular decomposition method for CPP using Boustrophedon patterns for complete area coverage within the cells of the decomposed polygonal area. The planner is implemented with two goals: (i) allowing for a choice of an arbitrary starting point and (ii) stitching together different coverage paths optimally. The requirement for these goals as well as the description of their implementation is given ahead:

\begin{figure}[!h]
      \centering
      \includegraphics[width=8.5cm]{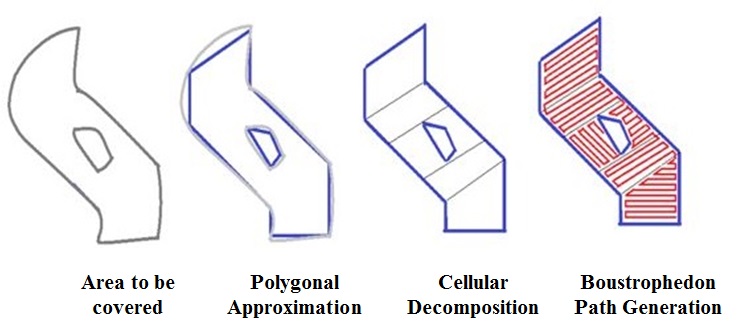}
      \caption{BCD algorithm implementation}
      \label{figurelabel}
   \end{figure}
\subsection{Exact Cellular Decomposition for CPP:}
Cellular decomposition methods for CPP can be utilized in a similar way as for general motion planning \cite{c19}. In this case, the polygonal area is decomposed into convex cells for which the Boustrophedon patterns can be separately implemented for complete coverage \cite{c20}. The description of the decomposition method for the implemented system is as follows: 
\smallskip

\subsubsection*{Boustrophedon Cellular Decomposition}
Choset and Pignon in 2001 \cite{c11} developed the Boustrophedon Cellular Decomposition (BCD) method for CPP which provided an effective way of dividing the polygon into convex cells for area coverage. This algorithm uses a split-and-merge technique at events defined by  vertices to generate these convex cells - the process is exemplified in Figure 2.

   \begin{figure}[!h]
      \centering
      \includegraphics[width=8.7cm]{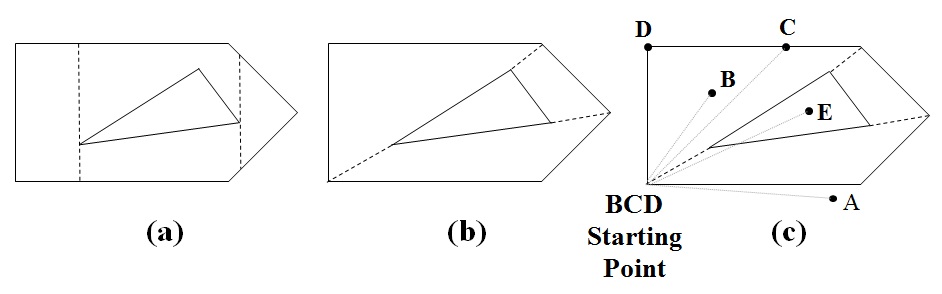}
      \caption{BCD for Coverage Planning - Drawbacks: (a) Cellular Decomposition with BCD approach, (b) Improved Decomposition with a Non-Unidirectional Sweeping Line, (c) Different Types of Initial Points }
      \label{figurelabel}
   \end{figure}      
From the final path in the Figure 2, it can be seen that the BCD algorithm uses a uni-directional sweeping line for decomposition which may result in a longer motion path than is required for coverage as shown in Figure 3 (b). Similarly, the choice of the initial point can also affect the length of the coverage path as shown in Figure 3 (c).

\subsubsection*{Path Length Optimality - Minimum-Sum-of-Altitudes Criterion}
The generic Boustrophedon Pattern for a convex polygon, shown in Figure 4, best describes its nature - it consists of parallel zig-zag patterns equally spaced from each other at a distance equal to the robot coverage radius with each line extending till the polygon boundary. (Here, the polygonal area is denoted with a bold upperface letter with its vertices denoted by the respective lowercase letter indexed as per the counter-clockwise sequence fed as input, e.g., $\textbf{P} = \{ p_1,p_2, \hdots, p_n\}$ - this notation is followed throughout the paper).
   \begin{figure}[!h]
      \centering
      \includegraphics[width=8.7cm]{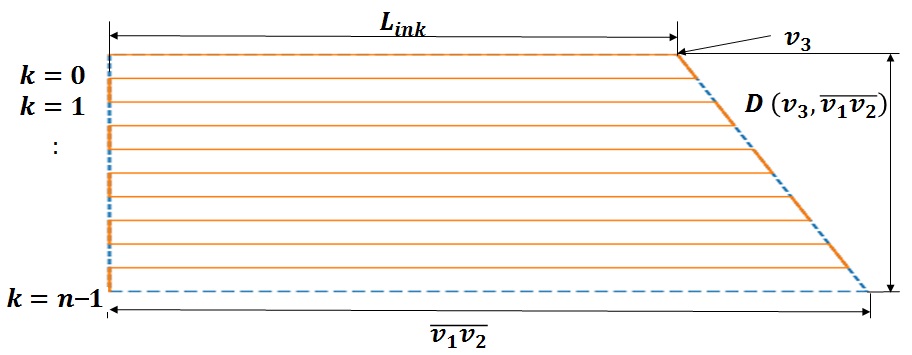}
      \caption{Boustrophedon Pattern for Coverage}
      \label{figurelabel}
   \end{figure}
   
Consider the pattern in Figure 4, where the total length of the Boustrophedon path $L_{tot}$, as expressed by Equation 1, describes it as a function of the sweep lengths, $L_{in,k}$ and the turn lengths $L_{turn,k}$.
Now, since the coverage radius $r$, is much smaller than the polygon width, minimizing $L_{tot}$ is possible through minimizing the number of turns, $n$ as shown in Equation 2.

This provides a metric of achieving distance optimality by simply minimizing the number of turns. Also, since the energy of turning is more than that of straight line motion, the above equation is also known to be energy optimal \cite{c2}.
The above derivation thus relates time-distance optimality for a coverage path to reducing the number of turns. This is more directly achieved through the Minimum Sum of Altitudes (MSA) criterion by Huang \cite{c8}, by finding the optimal sweep direction $\theta_{MSA}$ that solves the cost function in Equation 3  (see Figure 4 for notations).
\begin{eqnarray}
L_{tot}        & = & \sum\limits_{k=1}^n L_{in,k}+n\cdot L_{turn,k} \\
\min |L_{tot}| & = & \min\limits_n |\sum\limits_{k=1}^n L_{in,k}+n\cdot L_{turn,k}| \\
\theta_{MSA}   & = & \min\limits_i\{\>\>\max\limits_m\>\> [D(v_m,\overline{v_i v_{i+1}})\>\>]\>\>\}  \\
               & for &   \>\>     i  =  0,...,n-1; \nonumber\\
               &     &    m\neq i,i+1 \nonumber
\end{eqnarray}

Here, $D(.)$ represents the span defined by the edge $\overline{v_iv_{i+1}}$, i.e., the length of the longest altitude from the edge $\overline{v_iv_{i+1}}$ to a vertex $v_m$ while $\theta_{MSA}$ is the direction of the optimal edge. The above cost implies that for a convex polygon, the optimal direction giving the minimum number of turns is the direction of the edge that has the minimum sum of altitudes and therefore the minimum span \cite{c1}. Using this cost, it is thus possible to find the correct direction optimizing the number of turns and therefore the total path length as long as the polygon is convex.  

\subsubsection*{Non-Unidirectional Approach for Decomposition}
Figure 3 illustrates how the use of a Non-unidirectional sweeping line can result in shorter paths for the coverage pattern. The simplest way for such a decomposition is through polygon triangulation which decomposes an $n$-sided polygon into $n-2$ convex triangles using algorithms such as the Ear-Clipping algorithm \cite{c21}. The triangular cells can thus be merged into convex polygons wherever possible to form a reduced convex set - it is here that the MSA criterion is used that ensures the merged polygons remain optimal in terms of path length. The following cost is thus developed for convex decomposition:
\begin{align}
				 E_{MSA_i} & = \tau_{c}^{(ik)}\cdot[\>l_{m}(P_i) + l_{m}(P_k) - l_{m}(P^{m}_{ik})\>] & & & &\\
where \>\>\tau_{c}^{(i,k)} & =  1, \hspace{2mm} if \hspace{3mm} P^{m}_{ik} \hspace{5mm} \textnormal{is convex} \nonumber & & & &\\ 
				           & =  0, \hspace{2mm} if \hspace{3mm} P^{m}_{ik} \hspace{5mm} \textnormal{is concave} \nonumber & & & &
\end{align}

Here, the $l_{m}(.)$ is the length of the MSA optimal coverage path for the cell, $\tau_{conv}^{(ik)}$ is the convexity predicate, $P_i$ is the $i^{th}$ polygon and $P^{(m)}_{ik}$ is the merged polygon from $P_i$ and the $k^{th}$ adjoining polygon.
\smallskip

\subsubsection*{Choice of an Arbitrary Point}
As shown in Figure 3 (c), the initial point for coverage path can be of three types: outside the target area (Points A,E), inside the area (Point B) or on the boundary (Point C, D). For a point on the boundary, the problem can be solved using the Ear-Clipping algorithm during the convex decomposition - the polygon is rotated so that the vertex closest to the chosen point is the leftmost point. A similar case can be considered for an arbitrary point outside the polygon. For a point inside the polygon, choosing a vertex closest to the point may result in path overlap and requires an approach other than Boustrophedon patterns - such a case is not considered in this paper.
\smallskip

\subsubsection*{Convex Decomposition Algorithm}
The convex decomposition technique implemented for the offline planner is shown in Algorithm 1. As can be seen, the algorithm is derived from the Ear-Clipping Algorithm to include the MSA criterion so that the individual convex cells generated have the least possible length for the coverage path with regard to the starting point. 

The algorithm is initialized with the user-selected inputs: the input polygon, $P_{in}$ and the arbitrary initial point, $p_0$. The polygon then proceeds in a manner similar to  a Ear-Clipping algorithm which can be described as follows:

Step 1 for the algorithm selects the initial point for ear-clipping to be the point closest to the arbitrary initial point. For a generic Ear clipping algorithm, the adjacent vertices of the current chosen point are connected successively so that the polygon is partitioned into a disjoint set of triangles. To use this for convex decomposition, the algorithm can be modified to also recursively merge the triangles formed by ear-clipping into the prior cell if such a merged cell is found to be convex. In this manner, a convenient convex decomposition of a monotonic polygon can be carried out through the Ear-Clipping algorithm.

Now, it can be seen that steps 3-8 in Algorithm 1 follow the same process for the decomposition of the polygon as described above. However, Step 9 adds one more parameter to the algorithm which also calculates the MSA cost of the cells formed by splitting the polygon. For each iteration, the algorithm considers all the adjacent vertices of the current point, determines the MSA cost as well as the convexity of the cells formed by merging the ear-clipped triangles and selects the convex cell that gives the least MSA cost (Step 10). If this MSA cost is found to be larger than the sum of the MSA costs of the unmerged cells, then Steps 11-16 discard the merged polygon and proceed to the next triangle; otherwise the merged is selected to remain in the set of partitioned cells as the algorithm moves to the next point. These steps are iterated till all the vertices are covered.

The decision to merge a cell depending upon the convexity and MSA optimality makes sure that every convex cell within the partitioned polygon is MSA optimal.
         
\smallskip
\begin{algorithm}
\caption{Convex Decomposition Using MSA Criterion}\label{euclid}
\footnotesize
\textbf{Input:} \hspace{1mm} Polygon, \hspace{3mm} $\textbf{P}_{in} = (\textbf{p}_{in_i} : i=1,\hdots, n )$
  
\hspace{10mm} Initial Point, $\textbf{p}_{0}$

\textbf{Output:} Polygon Cells, $\textbf{C} = \{\textbf{C}_1, \hdots, \textbf{C}_m \}$ 
\smallskip
\begin{algorithmic}[1]
\State \textbf{Set} Initial Index, \hspace{2.5mm}$a^* \leftarrow \arg \min\limits_{i} || \textbf{p}_0 - \textbf{p}_{in_i}||$
\State \textbf{Set} Current Cell, \hspace{2mm}$\textbf{C}_{c} \leftarrow (\textbf{p}_{{in}_{a^*-1}},\textbf{p}_{{in}_{a^*}}, \textbf{p}_{{in}_{a^*+1}})$
\State \textbf{Set} New Polygon, \hspace{1mm}$\textbf{P}_{c} \leftarrow (\textbf{p}_{c_i} : i \neq a^*,\textbf{p}_{c_i}  \in \textbf{P}_{in} )$
\State \textbf{Set} Counter, \hspace{8mm} point$\_$ctr $\leftarrow 1$
\smallskip
\While{point$\_$ctr $\leq n-2$} 
\For{k \textbf{in} $(a^*,a^*-1)$ } 
\State $\textbf{C}_{new}(k) \leftarrow (p_{c,k-1},p_{c,k},p_{c,k+1})$
\State $\textbf{P}_{new}(k)\> \leftarrow$ Merge $\textbf{C}_{c}$ and $\textbf{C}_{new}(k)$
\State $\textbf{E}_{MSA}(k) \leftarrow$ Get MSA Cost for $\textbf{P}_{new}(k) $ 
\EndFor
\smallskip
\State $k^* \leftarrow \arg \min\limits_{k} \{\textbf{E}_{MSA}(k) : k \in (a^*,a^*-1)\}$
\smallskip
\If{$\textbf{E}_{MSA}(k^*) > 0$}
\State Remove $\textbf{C}_{c}$ from \textbf{C}
\State $\textbf{C}_{c} \leftarrow \textbf{P}_{new}(k^*)$
\Else 
\State $\textbf{C}_{c} \leftarrow \textbf{C}_{new}(k^*)$
\EndIf
\State Add $\textbf{C}_{c}$ to \textbf{C} 
\State $\textbf{P}_{c} \leftarrow (p_{c_i}: i \neq k^*, i = 1,\hdots, length(\textbf{P}_{c}) ) $
\State point$\_$ctr ++
\EndWhile
\end{algorithmic}
\end{algorithm}

\subsection{Generation of a Continuous Total Coverage Path}
While the above section describes distance-optimality of  coverage paths within individual Wells, the total coverage path is formed by stitching together these individual paths thus posing itself as a Travelling Salesman Problem. How this approach is modified for a shorter path is shown ahead:

\subsubsection{The Travelling Salesman Problem (TSP) Approach}
The treatment of total path generation problem as TSP is straightforward for the case of a cellular decomposition method. The centers of each cells are considered points of interest and a distance matrix is formed which, through dynamic programming, can find the shortest path joining all the cells \cite{c22}. However, for CPP, it is the start- and end-points of the cell coverage paths that need to be connected and not the centers which modifies the TSP approach accordingly.

\subsubsection{Modification of the TSP Problem}

As shown in Figure 5 (a), generation of an MSA optimal coverage path can allow for four possible pairs of start and end points - hence, the distance matrix for TSP formulation for an n-cell polygon is modified to form a $4n \times 4n$ matrix or an $n \times n$ matrix of $4 \times 4$ submatrices where each submatrix along a column corresponds to the respective cell of the polygon. Now, the submatrix in the first column is selected from the user-defined initial point - by MSA optimality, once the start point is selected, the end point that pairs with it is also selected (See Figure 5 (a)). This end point is then used in the next submatrix column to find the nearest start point for the next cell. This is repeated for all the columns till the cell are covered. The output paths generated by classic TSP and Modified TSP approach are shown in Figure 5 (b).
    
\section{ONLINE PLANNING - COVERAGE SLAM}

The online planner, like the offline planner, also needs to be modified to suit coverage path planning applications. This section elaborates these modifications as have been blockwise illustrated in Figure 1.
   \begin{figure}[!h]
      \centering
      \includegraphics[width=8.5cm]{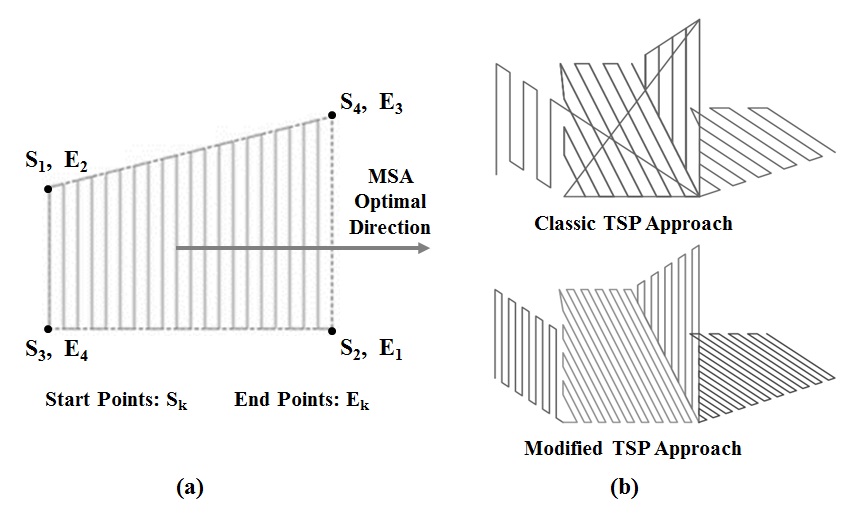}
      \caption{\footnotesize Total Path Generation for CPP: (a) Pairs of Start/End-points for an MSA Optimal Coverage Path (b) Total Paths generated with and without the modified TSP approach}
      \label{figurelabel}
    \end{figure}
    
\subsection{GMapping SLAM}
SLAM techniques allow for a robust estimation of robot position as well as the interest points in the map simultaneously through Bayesian inference. The system in Figure 1 makes use of GMapping SLAM method \cite{c18} in the online planner for SLAM operation - this technique uses particle filters \cite{c14} as the framework for pose and map estimation while 2D occupancy grids are used as features for map estimation. Use of occupancy grids have the added advantage of being inherently polygonal so that the currently observed occupancy grid can be dynamically checked with the initial map estimate for any map updates. 

\subsection{Dynamic Area Update:}
The offline planner generates the coverage path for an initial map estimate of the target area but does not allow for updating any path changes observed during navigation. When such a deviation from the initial map is observed, the online planner must choose between two actions: (i) complete traversal of the original coverage path resulting in partial coverage (ii) rerun the offline planner for the current map to give a new coverage path. 

Now, the first action is not favorable because if the current map detects new obstacles that can affect the path of the robot, it may result in failure of operation. On the other hand, if the offline planner re-initializes the coverage path for every small deviation, it will also result in an unstable behavior. Thus, it becomes necessary to quantify the change in the target region to make a measured decision of whether or not to re-initialize the coverage path.

\begin{figure}[!h]
      \centering
      \includegraphics[width=8.7cm]{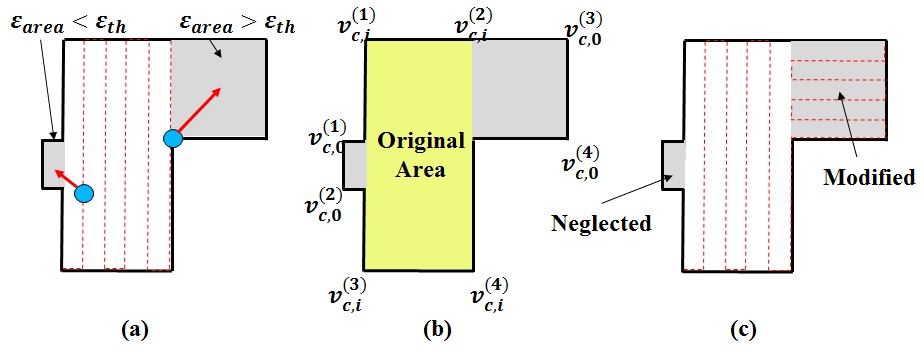}
      \caption{Dynamic Area Update by Online Planner}
      \label{figurelabel}
    \end{figure}

Figure 6 shows how the online planner tackles dynamic updates to maps as observed through SLAM. The polygonal boundary of the current occupancy grid $\textbf{O}_c$ is first extracted to obtain a set of vertex co-ordinates  $\textbf{V}_c = \{ \textbf{v}_{c}^{(k)} : k = 1, \hdots,m\}$ - these vertices are then associated with the original map using ICP registration \cite{c23} to create a set of inliers $V_{c,i} \{ \textbf{v}_{c,i}^{(k)} : k = 1, \hdots,p\}$ and outliers $V_{c,o}^{k} \{ \textbf{v}_{c,o}^{(i)} : i = 1, \hdots,(m-p)\}$. This is used to calculate the following cost:

\begin{equation}
\tilde{\textbf{E}}_{area} = |Area(\textbf{V}_c) - Area(\textbf{V}_{c,i})|
\end{equation}

As shown in Figure 6, the online planner can decide whether or not to re-initialize the coverage path depending upon the value of $\tilde{\textbf{E}}_{area}$. The threshold value $\epsilon_{th}$ is taken to be the product of the average length of sweeps in the polygon and the robot coverage radius. If the coverage path is re-initialized, the robot is required to begin path traversal from the current position at which the change was detected (it is here that the problem of an arbitrary initial point arises which is tackled in Section IV).

\subsection{Area Coverage Function}
The dynamic measure of system performance for a CPP problem requires the calculation of an area coverage function which measures the total area currently covered by the robot with regard to the target area. Assuming that the robot has a fixed coverage radius, the area covered as a function of time can be calculated from the robot path as follows:
\begin{equation}
A_{cov}(t) = \iint\limits_{P(x,y,0)}^{P(x,y,t)} R(x - \alpha, y- \beta) \cdot P(\alpha,\beta,t) \>\> d\alpha\ d\beta
\end{equation}

where $P(x,y,t)$ is the path traced by the robot at time t and $R(x,y)$ is the robot coverage area. For every control input to the robot, therefore, the incremental area covered $\delta A_{cov}$ corresponds to the difference between the area covered at the successive times. Now, with SLAM operation, $P(x,y,t)$ is replaced by its belief value \cite{c15} that is a spatially probabilistic. Thus, the value for $\delta A_{cov}(t)$ also peaks stochastically and can be determined for values $\delta x^*(t),\delta y^*(t) $ where Equation 6 is maximum - this determines the heading for the robot to maximize area coverage thus providing an appropriate area coverage function. This area coverage function allows a metric for CPP performance to be determined from the SLAM operation as long as the robot can track the history of its pose estimates (which basically implies a Full SLAM operation). As the robot pose is an outcome estimated by any SLAM framework, this allows the area coverage function to be independent of the underlying SLAM algorithm used for the online planner and can thus be utilized without affecting the SLAM performance.
 
\section{IMPLEMENTATION}
As a first stage in this phase of research, the proposed SLAM-assisted coverage path planning approach is implemented for an Unmanned Ground Vehicle (UGV) for indoor mobile mapping system. Roomba Create2 is utilized as UGV platform that is equipped with 2D LiDAR sensor (RPLiDAR-A2) as shown in Figure 7. Roomba Create2 is applied as a programmable robot that is easy to program and control by using several programmable kits (i.e., Raspberry pi, and Arduino kit). Due to its compact design, it is applicable to make some design modifications by constructing extra plates to install more sensors \cite{c24}.The RPLiDAR-A2 is a low cost 360-degree 2D LiDAR which can generate 4,000 pulses per second with high rotation speed as well as performing 360-degree scan within 6-meter range. This system is implemented in a ROS framework using GMapping SLAM. The system can also be extended for use with the 3D LiDAR, VLP-16, that has 16 radially-oriented laser rangefinders - for this system, the vertical FOV is from -15˚ to +15 ˚ and the horizontal FOV is 360˚ while the effective range is from 1 m to 100 m depending on the application, and the point capture rate is around 300,000 points per second \cite{c25}.
\begin{figure}[!h]
      \centering
      \includegraphics[width=8cm]{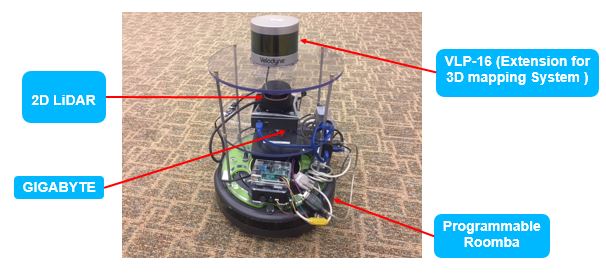}
      \caption{System Implementation using Roomba Create 2}
      \label{figurelabel}
    \end{figure}
    
\begin{figure*}[!h]
      \centering
      \includegraphics[width=17cm]{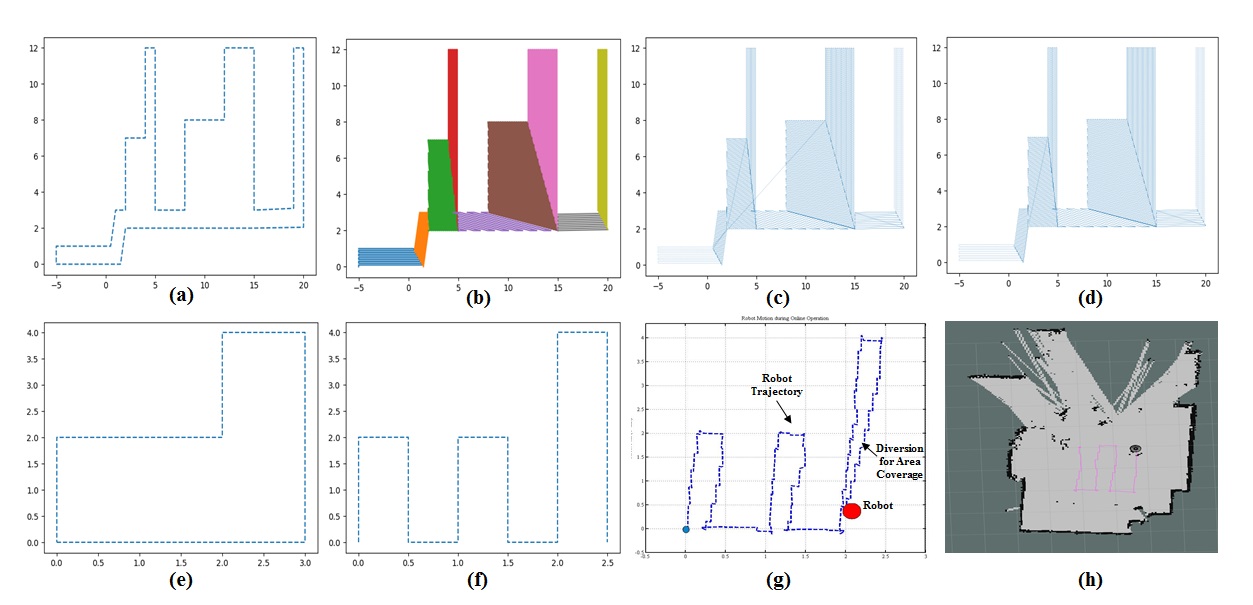}
      \caption{ Results for Offline Planner with Step-by-Step illustration: (a) Original Area, (b) Convex Decomposition with Boustrophedon Pattern (c) Total Coverage Path Generated (with Classic TSP), (d) Total Coverage Path Generated (with modified TSP). Results for Online and Offline Planner: (e) Original Area, (f) Total Coverage Path Generated, (g) Robot Path during Navigation, (h) Map Generation in RViz (ROS Indigo).}
      \label{figurelabel}
    \end{figure*}

\section{RESULTS}

The above implementation has been used for executing the online and offline planner for performance and analysis, the results of which are described in this section.

Figures 8 (a-d) shows the output of the offline planner with each of its stages illustrated: Convex Decomposition with Boustrophedon Path Generation and Total Path Generation. As can be seen, the final path generated with Modified TSP approach in Figure 8 (d) is more refined than the one without the modified approach in Figure 8 (c).
\begin{table}[h]
\caption{System Performance for Different Inputs}
\label{table_example}
\begin{center}
\begin{tabular}{|p{0.8cm}|p{0.8cm}|p{0.8cm}|p{0.9cm}|p{1.1cm}|p{1.1cm}|}
\hline
\begin{center} Input 						\end{center}& 
\begin{center} Number of Vertices 			\end{center}& 
\begin{center} Number of Turns (BCD) 		\end{center}& 
\begin{center} Number of Turns (New) 		\end{center}&
\begin{center} Path Length (TSP) 	(m)	\end{center}&
\begin{center} Path Length (New)	(m)	\end{center}\\
\hline
 & & & & & \\
Area 1 & 4  & 20 & 20 & 120.530   & 120.530\\
Area 2 & 5  & 31 & 26 & 196.565   & 189.616\\
Area 3 & 7  & 37 & 30 & 269.331   & 254.358\\
Area 4 & 11 & 61 & 59 & 1089.711  & 1006.794\\
Area 5 & 25 & 191& 183& 15143.076 & 14895.724\\
\hline
\end{tabular}
\end{center}
\end{table}

Figures 8 (f-h) shows the step-by-step operation of the combined offline and online planner on a target area with the implemented system. Figure 8 (h) shows the actual trajectory of the robot during online navigation. A loop closure error of $(3.1715 \>\>cm,0.1773 \>\>cm,1.71^{\circ})$ is observed from system performance - this value of the loop closure shows that the hybrid planning strategy affects the SLAM system minimally and can thus be possibly extended to large scale SLAM frameworks such as RTABMAP \cite{c19b}.  

Table 1 shows the system response to different input areas in comparison to unmodified methods - the modified approach is observed to provide better results in all cases. The polygonal areas 1-4 were randomly generated for an increasing number of vertices (polygonal area 5 corresponds to a 2D layout of a corridor) and the results are obtained for both classic and modifed TSP approaches. As can be seen, the total length for the new approach is always observed to be less than or equal to the classical approach - this reduction results from the fact that the new approach seeks to minimze the overlap due to the connecting paths between the different cells. This can also be visually verified from the outputs for Area 4 (Figure 5 (b)) and Area 5 (Figures 8(c),8(d)).

The system implemented above presents a successful hybrid planning strategy for robot navigation that can be deployed for autonomous surveying with SLAM-assisted frameworks. Our experiments with extraction of sparse planar point clouds from high density 3D LiDAR point clouds have allowed  us to extend 2D SLAM techniques for real-time use with 3D LiDARs with a reduced computational burden. This has established the possibility of using this hybrid strategy with 3D LiDARs for autonomous navigation - developing this framework and comparing it with existing LiDAR-based SLAM systems for computational efficiency and speed is a part of our ongoing research.

\section{Conclusion}
The system operation illustrated shows the efficiency of area coverage operation improves through the modification of offline and online strategies presented in the paper. The shorter path length with an acceptable SLAM performance shows a coalition of the two methods for operation - such an approach can also be proposed to be extended to other types of path planning methods in the near future.



%

\end{document}